\documentclass{article} 
\usepackage{iclr2026_conference}
\usepackage{times}

\usepackage{amsmath,amsfonts,bm}

\def\eqref#1{equation~\ref{#1}}

\def\1{\bm{1}}

\DeclareMathAlphabet{\mathsfit}{\encodingdefault}{\sfdefault}{m}{sl}
\SetMathAlphabet{\mathsfit}{bold}{\encodingdefault}{\sfdefault}{bx}{n}

\usepackage{hyperref}
\usepackage{url}
\usepackage{booktabs}
\usepackage{mathrsfs}
\usepackage{adjustbox}
\usepackage{multirow}
\usepackage{xcolor} 
\usepackage{makecell}
\usepackage{multicol}
\usepackage{tcolorbox}
\usepackage{changepage}
\usepackage{graphicx}

\title{From Denoising to Refining: A Corrective Framework for Vision-Language Diffusion Model}

\author{
\textbf{Yatai Ji\textsuperscript{1}},\quad
 \textbf{Teng Wang\textsuperscript{2}}\thanks{\ \ Corresponding author.},\quad
 \textbf{Yuying Ge\textsuperscript{2}}\footnotemark[1],
\\
 \textbf{Zhiheng Liu\textsuperscript{1}},\quad
 \textbf{Sidi Yang\textsuperscript{1}},\quad
 \textbf{Ying Shan\textsuperscript{2}},\quad
 \textbf{Ping Luo\textsuperscript{1}\footnotemark[1]}
\\
 \textsuperscript{1}The University of Hong Kong,\qquad
 \textsuperscript{2}ARC Lab, Tencent PCG
}

\iclrfinalcopy 
\begin{document}

\maketitle

\begin{abstract}

Discrete diffusion models have emerged as a promising direction for vision-language tasks, offering bidirectional context modeling and theoretical parallelization. However, their practical application is severely hindered by a train-inference discrepancy, which leads to catastrophic error cascades: initial token errors during parallel decoding pollute the generation context, triggering a chain reaction of compounding errors and leading to syntactic errors and semantic hallucinations. To address this fundamental challenge, we reframe the generation process from passive denoising to active refining. We introduce \textbf{ReDiff}, a \textbf{re}fining-enhanced \textbf{diff}usion framework that teaches the model to identify and correct its own errors. Our approach features a two-stage training process: first, we instill a foundational revision capability by training the model to revise synthetic errors; second, we implement a novel online self-correction loop where the model is explicitly trained to revise its own flawed drafts by learning from an expert's corrections. This mistake-driven learning endows the model with the crucial ability to revisit and refine its already generated output, effectively breaking the error cascade. Extensive experiments demonstrate that ReDiff significantly improves the coherence and factual accuracy of generated content, enabling stable and efficient parallel generation far superior to traditional denoising methods. 
Our codes and models are available at \textcolor{blue}{\url{https://rediff-hku.github.io/}}.
\end{abstract}

\section{Introduction}

Discrete diffusion models have recently emerged as a promising alternative to the dominant autoregressive (AR) paradigm for vision-language models (VLMs)~\citep{DBLP:journals/corr/abs-2505-16933/llada-v, DBLP:journals/corr/abs-2505-15809/mmada, DBLP:journals/corr/abs-2505-16839/lavida, DBLP:journals/corr/abs-2505-20147/fudoki, DBLP:journals/corr/abs-2503-20853/unidisc, DBLP:conf/cvpr/LiLSFKYW25/ddit, DBLP:journals/corr/abs-2505-16990/dimple}. Unlike AR models, which generate text token-by-token in a fixed unidirectional manner, diffusion models conceptualize generation as an iterative denoising process. This approach allows for bidirectional context modeling, granting them greater flexibility in controlling the generation process and a theoretical potential for massive parallelization, promising significant gains in inference efficiency~\citep{nie2025large/llada, ye2025dream, DBLP:journals/corr/abs-2508-02193/seed-diffusion, DBLP:journals/corr/abs-2505-22618/fast-dllm}.

\begin{figure*}[t]
	\centering
	\includegraphics[width=\linewidth]{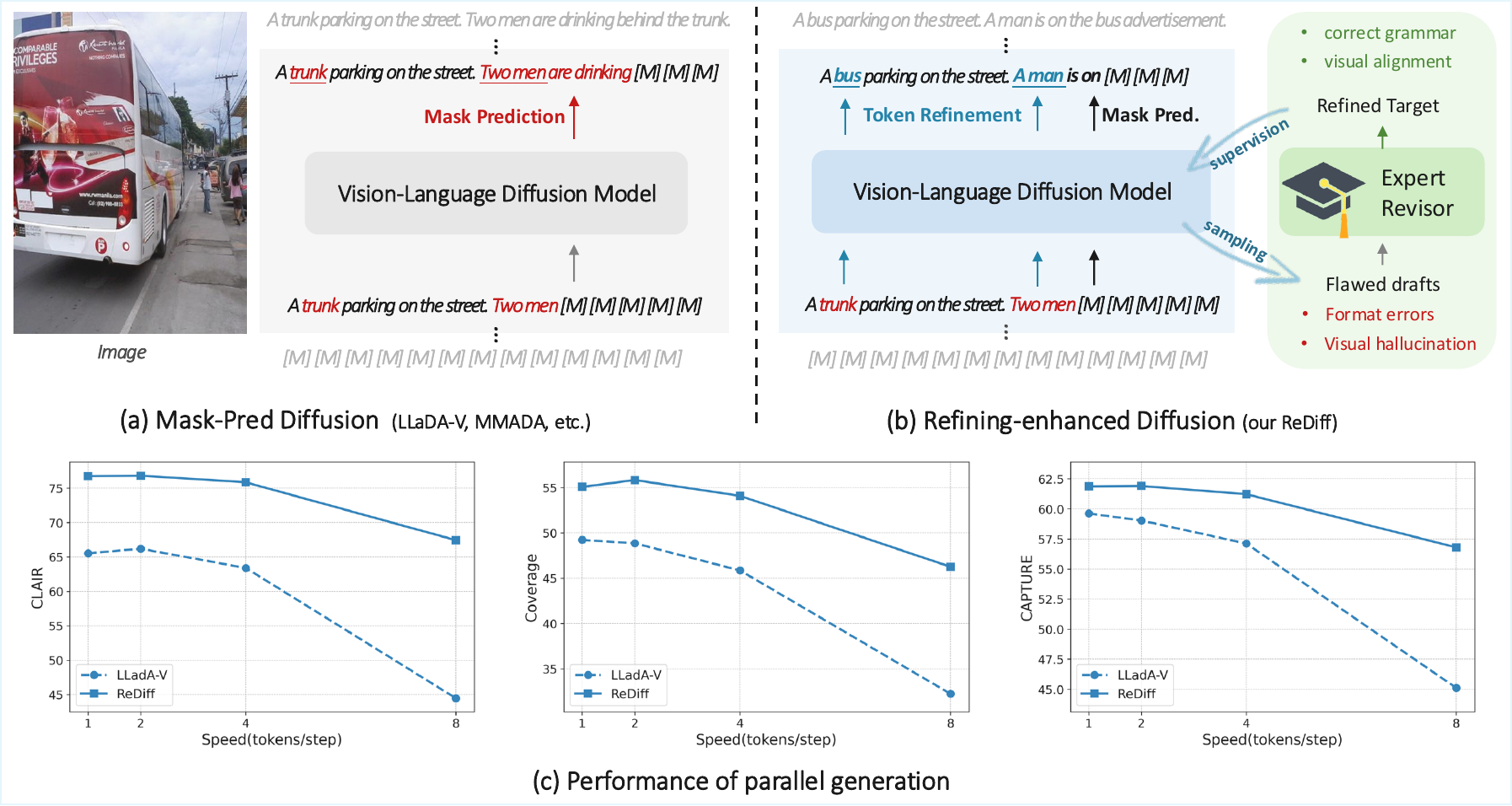}
    \vspace{-0.2cm}
	\caption{Comparison between standard vision-language diffusion models and our proposed refining-enhanced approach. (a) \textit{Mask-pred diffusion} is trained for passive denoising (mask recovering under fixed context). An initial error, such as misidentifying the ``bus" as a ``trunk", triggers an error cascade. The model cannot correct this mistake and proceeds to hallucinate further details based on the flawed context (e.g., ``Two men are drinking''), leading to a factually incorrect description. (b) \textit{Refining-enhanced diffusion} introduces a paradigm of active refining, teaching the model not only to predict masked tokens but also to perform token refinement. Our ReDiff learns to self-correct through an online loop where its own ``flawed drafts" are revised by an expert revisor. As a result, the model can identify and correct its initial mistakes (revising ``trunk" to ``bus", ``Two men" to ``A man"), breaking the error cascade and generating a factually grounded response. 
    (c) Performance comparison between LLaDA-V~\citep{DBLP:journals/corr/abs-2505-16933/llada-v} and ReDiff under different inference speeds. ``CLAIR'' and ``Coverage'' are detailed caption metrics on CapMAS~\citep{DBLP:journals/corr/abs-2412-15484/capmas}, and ``CAPTURE'' is on DetailCaps-4870~\citep{dong2024benchmarking/capture}. Our model delivers superior generation quality and achieves more stable results when using fewer inference steps.
    }
	\label{fig:teaser}
 \vspace{-0.3cm}
\end{figure*}

However, a significant gap exists between the theoretical promise and the practical reality of these models. Existing discrete diffusion models~\citep{nie2025large/llada, DBLP:journals/corr/abs-2505-16933/llada-v, DBLP:journals/corr/abs-2505-16839/lavida} are often plagued by incoherent and hallucinated artifacts (e.g., formatting errors like sequential commas or visually misaligned text) when parallel generation, frequently defaulting to one-token-per-step decoding process. We argue that these shortcomings are symptoms of a deeper, more fundamental problem: the error cascade driven by a training-inference discrepancy. Models are trained exclusively on clean, ground-truth data but are required at inference to generate from their own noisy, intermediate outputs. In a parallel decoding scenario, this discrepancy becomes catastrophic. As illustrated in Figure~\ref{fig:teaser}~(a), an error in a few tokens instantly pollutes the context for all other tokens being generated in parallel, initiating a cycle of compounding errors, which produces a detailed yet entirely fabricated description of the input image.

To break this vicious cycle, we propose a paradigm shift: from passive  denoising (mask recovering under fixed context) to active refining. We introduce a corrective framework for vision-language diffusion models, called ReDiff, which systematically teaches the model to identify and correct its own errors during denoising. 
Unlike previous models that merely fill masked tokens, ReDiff actively refines the entire context to guide the generation process. 
Our approach consists of a two-stage training process. First, we instill a foundational revision capability by training the model to correct synthetic errors, such as random token corruptions and injected hallucinations, moving beyond simple denoising to build a general capacity for revision. Second, we introduce an online self-correction loop where the model is forced to confront and learn from its own mistakes. By capturing its flawed``drafts" during training and learning to predict an expert's revision, the model directly mitigates the training-inference discrepancy.

This mistake-driven learning endows the model with a crucial, previously absent capability: the ability to revisit and refine its own outputs, including previously unmasked tokens. By learning to self-correct, our model develops robustness to its own imperfections, effectively breaking the error cascade and enabling robust parallel generation. As shown in Figure~\ref{fig:teaser}~(b), our refinement-based model successfully identifies and revises an initial error, leading to a more factually grounded and accurate generation. Our contributions are threefold:

1) We propose a new perspective that reframes the generation process of diffusion models from passive denoising to active, iterative refining to address the core challenge of error cascades.

2) We design and implement a two-stage training framework, featuring a core online self-correction loop that enables the model to learn to fix its own intrinsic errors.

3) Extensive experiments demonstrate that our method significantly improves the coherence and factual accuracy of generated content, exhibiting stability far superior to traditional denoising methods, especially in challenging few-step parallel generation scenarios, thereby greatly enhancing inference efficiency.
%

\section{Related Work}

\subsection{Large Language Diffusion Models}

Discrete diffusion models~\citep{DBLP:conf/nips/AustinJHTB21/d3pm, DBLP:conf/icml/LouME24/sedd, DBLP:journals/corr/abs-2505-10446/rldiffusion, DBLP:conf/iclr/ArriolaGCYQHSK25/block, Sun2022ScorebasedCD, sahoo2024simple-diff} represent a class of generative models tailored for discrete data like text. In contrast to image diffusion models, which corrupt data by adding Gaussian noise towards a standard Gaussian prior, text diffusion models typically operate by replacing original tokens to degrade semantic content. Early approaches, such as D3PM~\citep{DBLP:conf/nips/AustinJHTB21/d3pm}, employed discrete Markov chains where a transition matrix is progressively applied to the input, corrupting it towards a uniform distribution (i.e., any token becomes any other with equal probability) or an absorbing state (e.g., a [MASK] token). More recently, mask-and-pred diffusion models have demonstrated significant empirical success. For instance, LLaDA~\citep{nie2025large/llada} achieves performance comparable to autoregressive large language models by generating sentences from a fully masked sequence, progressively unmasking tokens with the highest confidence. Similarly, Dream~\citep{ye2025dream} has shown strong results by initializing its parameters from a pre-trained autoregressive model.

Theoretically, discrete diffusion models offer advantages over traditional autoregressive models~\citep{DBLP:journals/corr/abs-2302-13971/llama, qwen3blog, bi2024deepseek-llm, vicuna, chatgpt}. Their bidirectional context modeling enables flexible and controllable generation, while their inherent parallelism promises significant acceleration in sampling speed. However, this potential for parallel generation remains largely untapped. Current models often suffer from output degradation—such as repetition and grammatical errors—when attempting to predict multiple tokens per step. Our work directly addresses this by enhancing the stability of parallel decoding. This aligns with a recent line of work exploring the correction of generated content. For example, SEED-Diffusion~\citep{DBLP:journals/corr/abs-2508-02193/seed-diffusion} introduced an ``edit-based forward process" for code generation, which adds edit-specific noise in the final 20\% of steps to allow for revisions. Likewise, FUDOKI~\citep{DBLP:journals/corr/abs-2505-20147/fudoki}, a multimodal model based on discrete flow matching, progressively revises a random sentence, where each word is uniformly sampled from the vocabulary, to the correct answer. Our method is distinct in that it treats revision not as another form of noise, but as a high-level refinement process. Specifically, our framework trains the model to learn from and correct its own characteristic errors, moving beyond simple noise reversal.

\subsection{Large Vision Language Models}

Large vision language models (LVLMs)~\citep{DBLP:conf/nips/LiuLWL23a/llava, DBLP:conf/nips/Dai0LTZW0FH23/instructblip, DBLP:journals/corr/abs-2408-03326/llava_ov, DBLP:journals/corr/abs-2308-12966/qwen, DBLP:conf/cvpr/JiTJKCZWY023/scl, wang2025internvideo2/internvideo} have achieved remarkable success in vision understanding and have been applied to a myriad of real-world scenarios~\citep{DBLP:conf/iclr/JiZWSC0YY025/ida-vlm, DBLP:journals/corr/abs-2307-03601/gpt4roi, DBLP:conf/nips/ChengYFGYK0L24/srgpt}. The dominant architecture connects a pre-trained vision encoder~\citep{radford2021learning/clip,tschannen2025siglip} to an autoregressive language model via a lightweight module like an MLP or Q-Former. These models first realize cross-model alignment with pre-training and then conduct visual instruction tuning to handle a wide range of vision-centric tasks.

Despite their success, a persistent challenge in LVLMs is the phenomenon of hallucination~\citep{bai2024hallucination}, where the model generates text that is factually inconsistent with the visual input. In autoregressive models, this issue is exacerbated by error propagation; an incorrectly generated token can irreversibly misguide the subsequent generation path. 
Notably, current multimodal discrete diffusion models, such as LLaDA-V~\citep{DBLP:journals/corr/abs-2505-16933/llada-v}, LaViDa~\citep{DBLP:journals/corr/abs-2505-16839/lavida}, and MMaDA~\citep{DBLP:journals/corr/abs-2505-15809/mmada}, also adhere to this limitation, fixing tokens in place once they are unmasked. Our ReDiff, however, leverages the bidirectional attention mechanism inherent to the diffusion paradigm. This allows our model to revisit and optimize already-generated content, enabling a progressive refinement process that directly mitigates hallucination.
\section{Methodology}

In this section, we introduce our refining-enhanced diffusion framework, ReDiff, designed to enhance the generation accuracy and stability of vision-language diffusion models. In contrast to traditional approaches that focus on recovering text from all [MASK] noise, our work emphasizes the high-level refinement of generated text. Guided by an expert model, our framework enables the model to learn from its own generation errors. This fosters a self-correction capability during inference, allowing it to simultaneously unmask new tokens while refining previously generated ones, thereby mitigating the problem of error cascades in parallel generation.

We will first present the preliminaries of discrete diffusion models in Section~\ref{sec:Preliminaries}. We then introduce the first stage of our approach, foundational revision training, in Section~\ref{sec:revision} . Section~\ref{sec:refine} details the core of our framework, online self-correction learning.  Section~\ref{sec:inference} details the inference process.

\begin{figure*}[t]
	\centering
	\includegraphics[width=\linewidth]{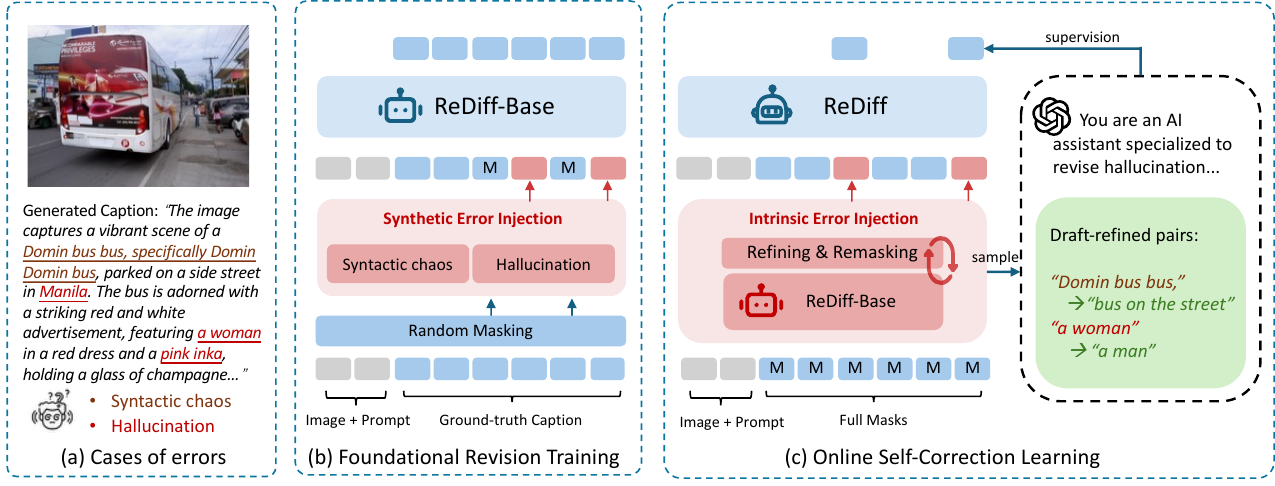}
	\caption{Overview of our proposed two-stage training framework for corrective refining. (a) We illustrate common failure modes in standard vision-language diffusion models, which are prone to generating syntactic errors (e.g., ``Domin bus bus") and factual hallucinations (e.g., ``a woman"). (b) In the foundational revision training stage, we instill a general corrective capability by training a base model (ReDiff-Base) to revise synthetic errors that are intentionally injected into ground-truth captions. (c) For the second stage, i.e., online self-correction learning, the model generates its own flawed ``drafts". These drafts, containing the model's intrinsic errors, are then revised by an expert AI assistant. The resulting ``draft-refined pairs" provide strong supervision, teaching our final model (ReDiff) to identify and correct its own characteristic mistakes, thus breaking the error cascade.}
	\label{fig:method}
\end{figure*}

\subsection{Preliminaries of Discrete Diffusion Models}
\label{sec:Preliminaries}

A discrete diffusion model formalizes text generation through a forward and a reverse process. The forward process gradually corrupts a clean text sequence $x_0$ into a noisy state $x_t$ over a series of timesteps $t\in[0,1]$. In mask-pred models, this is achieved by replacing tokens with a [MASK] token based on a noise schedule $\gamma_t$, culminating in a fully masked sequence as a prior distribution. The forward process is formulated as:
\begin{equation}
q \bigl(x_t[i]=c \,\big|\, x_0[i]\bigr)
=
\begin{cases}
1-\gamma_t, & \text{if } c = x_0[i],\\[0.5em]
\gamma_t,   & \text{if } c = \mathbf{M}.
\end{cases}
\end{equation}

The reverse process aims to reverse this corruption. Starting from a fully masked sequence, the model iteratively predicts the original tokens. At each step, it predicts probabilities for all masked positions, unmasks a few high-confidence tokens, re-masks the rest, and feeds the updated sequence back into the model for the next iteration.

The model, a parametric mask predictor, is trained to predict all masked tokens (denoted by a set $\mathbf{M}$) simultaneously. The training objective is a cross-entropy loss computed only on the masked tokens:
\begin{equation}
\mathcal{L}_{\text{CE}}(\theta) = -\mathbb{E}_{t, v, p_0, r_0, r_t} \left[ \frac{1}{t} \sum_{i=1}^{L_{r_0}} \mathbf{1}[r_t^i = \mathbf{M}] \log p_\theta(r_0^i | v, p_0, r_t) \right],
\label{eq:diffusion}
\end{equation}

where $v$ and $p_0$ denote visual content and prompt, $r_0$ is the correct response, $t$ is sampled uniformly, and $r_t$ is sampled from the forward process.

A key advantage of discrete diffusion models is their potential for parallel generation, where multiple tokens are unmasked in a single step, significantly reducing the number of required iterations. However, existing model~\citep{nie2025large/llada, ye2025dream, DBLP:journals/corr/abs-2505-16933/llada-v} treat already-unmasked tokens as fixed conditions for future predictions. If an incorrect token is generated, it can derail subsequent steps, leading to an error cascade. Yet, unlike the unidirectional attention in AR models, the bidirectional attention mechanism inherent to these models provides the architectural foundation for updating previously generated tokens, a potential we exploit in our framework.

\subsection{Stage I: Foundational Revision Training}
\label{sec:revision}

Observations of existing vision-language diffusion models, especially in few-step generation scenarios, reveal two predominant error types: syntactic chaos (e.g., incoherence, repetition, grammatical errors) and semantic hallucinations (content that contradicts the visual input), as shown in Figure~\ref{fig:method}~(a). In this first training stage, we teach the model to correct these two types of errors, extending its capability from simple denoising to foundational text revision.

We use two data construction ways. For syntactic errors, we corrupt the text from ground-truth image-text pairs by randomly replacing a fraction of tokens with other tokens from the vocabulary, creating syntactically chaotic inputs. 
For hallucination errors, we leverage pairs of correct captions and human-corrputed captions with factual errors (e.g., incorrect objects, attributes, or counts), which directly provide examples of visually inconsistent text.

As shown in Figure~\ref{fig:method}~(b), we task the model with restoring a ``polluted" response $r_t$ to its original, correct version $r_0$. We first apply the standard masking process to $r_0$ according to a sampled noise level $t$. Then, on the remaining unmasked tokens, we inject the synthetic errors described above. This corrupted sequence serves as the model's input. The model is trained to predict the entire original text $r_0$. The loss is computed not only on the [MASK] tokens but also on the syntactically corrupted tokens ($\mathcal{L}_{\rm syntax}$) and hallucinated tokens ($\mathcal{L}_{\rm hallucination}$). We also include a loss on the uncorrupted tokens ($\mathcal{L}_{\rm clean}$) to encourage the model to preserve correct content. The loss of each type is calculated as follows:
\begin{equation}
    \label{eq:revise}
    \mathcal{L}_{\text{type}}(\theta) = -\mathbb{E}_{t, v, p_0, r_0, r_t} \left[ \frac{1}{t} \frac{1}{N_{\text{type}}} \sum_{i=1}^{L_{r_0}} \mathbf{1}[r_t^i \in \text{type}] \log p_\theta(r_0^i | v, p_0, r_t) \right], 
\end{equation}

where $\rm type \in \{mask, syntax, hallucination, clean\}$. Each loss component is normalized by the number of its corresponding tokens $N_{\text{type}}$ to balance their contributions. The total loss is:
\begin{equation}
    \mathcal{L}_{\text{revision}} = \mathcal{L}_{\text{mask}} + \mathcal{L}_{\text{syntax}} + \mathcal{L}_{\text{hallucination}} + \mathcal{L}_{\text{clean}}.
    \label{eq:total}
\end{equation}

After Stage I, we obtain ReDiff-Base, a model equipped with the foundational capability to correct both syntactic errors and factual hallucinations. However, this stage has a limitation: the errors are synthetic and may not reflect the characteristic mistakes the model itself is prone to making.

\subsection{Stahe II: Online Self-Correction Learning}
\label{sec:refine}

To teach the model to fix its own idiosyncratic errors, we introduce an online self-correction learning framework. The process, illustrated in Figure \ref{fig:method} (c), proceeds as follows:
(1) Generating drafts: We use ReDiff-Base to generate a response for an image, denoted as $r_{\rm draft}$. We typically use decoding results of different generation steps to cover more mistakes. (2) Expert revision: The image $I$, the generated draft $r_{\rm draft}$, and the ground truth are fed to a powerful external ``expert model" (e.g., o4-mini). With a carefully designed prompt, the expert model identifies and corrects both grammatical and hallucinatory errors in $r_{\rm draft}$, producing a refined version, $r_{\rm refined}$. We specifically extract the pairs of erroneous and corrected segments. (3) Learning to refine: We form a new training instance $⟨I, r_{\rm draft}, r_{\rm refined}⟩$ and fine-tune our model on these data. Note that the training loss is computed only on the segments that the expert model identified and corrected. This targeted learning prevents the model from being penalized for other potential errors in the draft that the expert may have missed. The training loss is:
\begin{equation}
    \label{eq:refine}
    \mathcal{L}_{\text{refine}}(\theta) = -\mathbb{E}_{t, v, p_0, r_\text{draft}, r_\text{refined}} \left[\frac{1}{N_{\text{mistake}}} \sum_{i=1}^{L_{r_0}} \mathbf{1}[r_\text{draft}^i \in \text{mistake}] \log p_\theta(r_\text{refined}^i | v, p_0, r_\text{draft}) \right].
\end{equation}

To maintain the foundational capabilities learned in the first stage, we mix in a small amount of the Stage I data during this phase. This entire cycle can be iterated: the refined model from one round can be used to generate new drafts for the next round of expert revision and fine-tuning, progressively enhancing its self-correction ability. The key advantage here is that the model learns from its own mistakes, which is a more targeted and efficient way to improve its robustness and the stability of parallel generation.

\subsection{Inference Process}
\label{sec:inference}

Our inference process differs from that of traditional discrete diffusion models by integrating refinement into each generation step. Specifically, the process starts with a fully masked sequence. At each step, the model computes the output probability distribution over the entire vocabulary for all token positions. 
For masked positions, if the inference speed is $n$ tokens per step, we select the top-$n$ most confident tokens and unmask them. For previously unmasked positions, we replace the existing tokens with the newly predicted ones. 
This allows for the simultaneous unmasking of new content and refining of existing content. As more context is generated, previously generated tokens are iteratively updated to be more coherent and factually accurate, effectively reducing the occurrence of syntactic chaos and hallucinations.
\section{Experiments}

\begin{table}[]
\centering
\small
\caption{Performance comparison with state-of-the-art models on three detailed image caption benchmarks. The best scores of vision-language diffusion models are in \textbf{bold}.}
\label{tab:main}
\begin{adjustbox}{max width=\textwidth}
\begin{tabular}{l|ccc|c|c}
\toprule
\multicolumn{1}{c|}{\multirow{2}{*}{Model}} & \multicolumn{3}{c|}{CapMAS}                                                                & \multicolumn{1}{c|}{CapArena}      & \multicolumn{1}{c}{DetailCaps-4870} \\
\multicolumn{1}{c|}{}                       & \multicolumn{1}{c}{CLAIR} & \multicolumn{1}{c}{Coverage} & \multicolumn{1}{c|}{Factuality} & \multicolumn{1}{c|}{CapArena-Auto} & \multicolumn{1}{c}{CAPTURE}         \\ \midrule
\textit{\textbf{AR model}}                           &                                &                              &                            &                                    &                                     \\
LLaVA-1.5-7B~\citep{liu2024improved/llava15}                                 &  62.10  & 34.30  & 52.80         & -94.00           &              51.08                       \\
InternVL-2.5-7B~\citep{chen2024expanding/internvl25}       &          78.37 & 52.57 & 78.69        & -29.83   &                      57.80          \\
Qwen2.5-VL-7B~\cite{DBLP:journals/corr/abs-2308-12966/qwen}     &           80.48 & 57.32 & 82.73        & -16.83   & 60.61                               \\ \midrule
\textit{\textbf{Discrete diffusion model}}           &                                &                              &                            &                                    &                                     \\
MMaDA~\citep{DBLP:journals/corr/abs-2505-15809/mmada}                     &            35.45                    &                 14.33             &               57.98             &                        -97.00            &           19.55                          \\
FUDOKI~\citep{DBLP:journals/corr/abs-2505-20147/fudoki}                     &            51.94                    &              39.18                &          46.04                  &                     -98.83               &               57.92                      \\
LaViDa~\citep{DBLP:journals/corr/abs-2505-16839/lavida}                 &       56.22 & 44.18 & 53.57        & -90.00     &                 57.28                    \\
LLaDA-V~\citep{DBLP:journals/corr/abs-2505-16933/llada-v}                                     &           65.54                     &          49.22                    &               61.06             &                -77.17                  &                    59.62                 \\
ReDiff (ours)                                  &                    \textbf{76.74}            &             \textbf{55.07}                 &               \textbf{63.29}          &                  \textbf{-51.50}                 &                 \textbf{61.88}                    \\ \bottomrule
\end{tabular}
\end{adjustbox}
\vspace{-0.3cm}
\end{table}

\subsection{Experiment Settings}

\textbf{Training Setup.} 
Our primary focus is on enhancing the generative capabilities of vision-language diffusion models. We select detailed image captioning as the representative task to validate our framework, although the methodology is generalizable to other generative tasks. Our model is built upon the existing LLaDA-V model, leveraging its foundational mask prediction capabilities while endowing it with the ability to refine. The training data comprises caption datasets from LLaVA-1.5~\citep{DBLP:conf/nips/LiuLWL23a/llava}, ShareGPT4V~\citep{DBLP:journals/corr/abs-2311-12793/sharegpt4v}, and the ViCrit dataset~\citep{DBLP:journals/corr/abs-2506-10128/vicrit}. 
When constructing hallucination revision data in Stage I, we leverage the existing hallucination dataset ViCrit, which contains pairs of correct and hallucinated descriptions. 
For Stage I (foundational revision training), we use a total of 260k image-text pairs, with a random token replacement probability of 0.1 for creating syntactic chaos. For Stage II (online self-correction learning), we generate approximately 10k draft-refined caption pairs in each round. The drafts are generated with 128, 32, and 16 inference steps, and o4-mini serves as the expert model for revisions. The prompt for o4-mini is detailed in Appendix \ref{app:prompt}. Our experiments revealed that a single round of this online refinement training yielded the most significant improvements.

\textbf{Benchmarks and Evaluation Setup.} 
We evaluate our model on three recent benchmarks for detailed image caption: 
CapMAS~\citep{DBLP:journals/corr/abs-2412-15484/capmas} uses three metrics evaluated by GPT-4o: CLAIR for overall caption quality, Coverage for the comprehensiveness of the description, and Factuality for the accuracy of the content.  
CapArena~\citep{cheng2025caparena} employs a pairwise comparison methodology where the outputs of the test model are compared against those of three baseline models with GPT-4o. A final score is calculated based on these win ratio. 
DetailCaps-4870~\citep{dong2024benchmarking/capture} uses the CAPTURE metric, which scores the generated caption by comparing its scene graph to that of the ground-truth description. 
We compare ReDiff against several vision-language diffusion models, including LLaDA-V, LaViDa, MMaDA, and FUDOKI. We also include results from some typical AR-based VLMs. At inference, the maximum generation length is 128. An inference process of 128 steps corresponds to a speed of 1 token/step, while 32 steps correspond to 4 tokens/step.

\subsection{Main Results}

As shown in Table \ref{tab:main}, our ReDiff achieves state-of-the-art results among all diffusion-based models across each metric. On CapMas, our model's CLAIR score shows a remarkable 11.2 point improvement over the LLaDA-V, reaching a comparable level to InternVL-2.5. The Coverage and Factuality scores also increase by 5.85 and 2.23 points, respectively, indicating that our captions are not only richer in content but also more accurate. On CapArena, our model outperforms LLaDA-V by 25.67 points. Furthermore, we achieve a CAPTURE score of 61.88, surpassing the powerful Qwen2.5-VL. These results demonstrate that our refining-enhanced diffusion method effectively improves fluency and mitigates hallucinations, leading to a substantial enhancement in overall caption quality.

In Tables \ref{tad:step capmas} and \ref{tab:step capa}, we compare models trained with the traditional mask-pred objective versus our refinement framework, using identical datasets and base model. Our model consistently outperforms the mask-trained baseline at every step count. Crucially, as the generation speed increases (i.e., fewer steps), our model's performance degrades much more gracefully, demonstrating superior stability in parallel generation. For instance, on the CLAIR metric, as the speed increases from 1 token/step to 8 tokens/step, the mask-trained model's score plummets from 74.53 to 46.38, whereas our model's score only decreases from 76.74 to 67.44. Notably, our model's performance at 4 tokens/step is higher than that of both LLaDA-V and the mask-trained baseline at 1 token/step. A similar trend is observed for Coverage. The trend for Factuality is less pronounced, as the baseline's score does not drop significantly at fewer steps. This is because the metric relies on extracting valid items for verification; as the baseline's output becomes more chaotic, fewer items can be extracted, artificially stabilizing the correctness ratio. On both CapArena and CAPTURE, our model also demonstrates more robust parallel generation, with the CAPTURE score dropping by only 0.65 points when accelerating from 1 to 4 tokens/step.

\begin{table}[t]
\centering
\caption{Performance comparison of different inference steps on CapMas benchmark. ``Mask-pred training" indicates training with the traditional mask-pred objective using identical datasets.}
\label{tad:step capmas}
\begin{adjustbox}{max width=\textwidth}
\begin{tabular}{l|cccc|cccc|cccc}
\toprule
Metrics           & \multicolumn{4}{c|}{CLAIR} & \multicolumn{4}{c|}{Coverage} & \multicolumn{4}{c}{Factuality} \\
Speed (token/step) & 1       & 2  & 4      & 8  & 1       & 2   & 4       & 8   & 1        & 2   & 4       & 8   \\ \midrule
LLada-V           & 65.54   &  66.20  & 63.40   &  44.47  & 49.22   &   48.85  & 45.85   &   32.24  & 61.06    &   \textbf{61.10}  & 60.69   &   64.97  \\
Mask-pred training     & 74.53   &  73.57  & 69.23  &  46.38  & 54.15   &  54.11   & 47.60   &   29.69  & 59.68    &   58.43  & 59.66   &  \textbf{67.79}   \\
ReDiff        & \textbf{76.74}   &   \textbf{76.81} & \textbf{75.85}  &  \textbf{67.44}  & \textbf{55.07}   &  \textbf{55.82}   & \textbf{54.08}   &   \textbf{46.25}  & \textbf{63.29}    &  60.95   & \textbf{60.87}   &  65.14   \\ \bottomrule
\end{tabular}
\end{adjustbox}
\end{table}

\begin{table}[]
\vspace{-0.2cm}
\centering
\small
\caption{Performance comparison of different inference steps on CapArena and CAPTURE metrics.}
\label{tab:step capa}
\begin{adjustbox}{max width=\textwidth}
\begin{tabular}{l|cccc|cccc}
\toprule
Metrics           & \multicolumn{4}{c|}{CapArena-Auto} & \multicolumn{4}{c}{CAPTURE}   \\
Speed (token/step) & 1       & 2      & 4      & 8      & 1     & 2     & 4     & 8     \\ \midrule
LLada-V           & -77.17  & -84.00 & -90.50 & -99.00 & 59.62  & 59.04   &  57.12   &  45.11    \\
mask training     & -56.00  & -70.50 & -90.33 & -98.33 & 59.98 & 59.61 & 56.99 & 45.12 \\
ReDiff        & \textbf{-51.50}  & \textbf{-56.83} & \textbf{-72.67} & \textbf{-91.67} & \textbf{61.88} & \textbf{61.91} & \textbf{61.23} & \textbf{56.80} \\ \bottomrule
\end{tabular}
\end{adjustbox}
\end{table}

\begin{table}[t]
\centering
\small
\caption{Effect of each training stage in the refining-enhanced diffusion paradigm.}
\label{tab:train stage}
\begin{adjustbox}{max width=\textwidth}
\begin{tabular}{l|cc|cc|cc|cc}
\toprule
Metrics                & \multicolumn{2}{c|}{CLAIR}     & \multicolumn{2}{c|}{Coverage}  & \multicolumn{2}{c|}{Factuality} & \multicolumn{2}{c}{CapArena-Auto} \\
Speed (token/step)      & 1     & \multicolumn{1}{c|}{4} & 1     & \multicolumn{1}{c|}{4} & 1      & \multicolumn{1}{c|}{4} & 1               & 4               \\ \midrule
LLada-V                & 65.54 & 63.40                  & 49.22 & 45.85                  & 61.06  & 60.69                  & -77.17          & -90.50          \\
Base + Stage I  & 71.31 & 71.67                  & 51.73 & 51.83                  & 58.04  & 55.22                  & -69.17          & -73.17          \\
Base + Stage II & 73.02 & 73.52                  & 53.44 & 53.00                  & 59.49  & 57.40                  & -68.00          & -77.67          \\
Stage I + Stage II    & \textbf{76.74} & \textbf{75.85}                  & \textbf{55.07} & \textbf{54.08}                  & \textbf{63.29}  & \textbf{60.87}                  & \textbf{-51.50}          & \textbf{-72.67}          \\ \bottomrule
\end{tabular}
\end{adjustbox}
\vspace{-0.2cm}
\end{table}

\begin{table}[]
\centering
\small
\caption{Effect of different settings in the foundational revision training stage.}
\label{tab:stage1}
\begin{adjustbox}{max width=\textwidth}
\begin{tabular}{l|cc|cc|cc|cc}
\toprule
Metrics                & \multicolumn{2}{c|}{CLAIR}     & \multicolumn{2}{c|}{Coverage}  & \multicolumn{2}{c|}{Factuality} & \multicolumn{2}{c}{CapArena-Auto} \\
Speed (token/step)      & 1     & \multicolumn{1}{c|}{4} & 1     & \multicolumn{1}{c|}{4} & 1      & \multicolumn{1}{c|}{4} & 1               & 4               \\ \midrule
LLada-V                & 65.54 & 63.40                  & 49.22 & 45.85                  & \textbf{61.06}  & \textbf{60.69}                  & -77.17          & -90.50          \\
Revise hallucination   & 69.33 & 67.01                  & 51.08 & 46.61                  & 59.46  & 57.06                  & -74.33          & -87.67          \\
Revise syntactic errors      & 69.48 & 70.30                  & \textbf{52.12} & 49.96                  & 56.57  & 56.15                  & -69.67          & -88.83          \\
Dynamic revise ratio   & 68.26 & 67.98                  & 50.49 & 48.66                  & 59.23  & 56.60                  & -74.83          & -82.50          \\
Ours (ReDiff-Base) & \textbf{71.31} & \textbf{71.67}                  & 51.73 & \textbf{51.83}                  & 58.04  & 55.22                  & \textbf{-69.17}          & \textbf{-73.17}          \\ \bottomrule
\end{tabular}
\end{adjustbox}
\vspace{-0.2cm}
\end{table}

\begin{table}[]
\centering
\small
\caption{Effect of online self-correction learning rounds.}
\label{tab:stage-2}
\begin{adjustbox}{max width=\textwidth}
\begin{tabular}{l|cc|cc|cc|cc}
\toprule
Metrics                            & \multicolumn{2}{c|}{CLAIR}     & \multicolumn{2}{c|}{Coverage}  & \multicolumn{2}{c|}{Factuality} & \multicolumn{2}{c}{CapArena-Auto} \\
Speed (token/step)                 & 1     & \multicolumn{1}{c|}{4} & 1     & \multicolumn{1}{c|}{4} & 1      & \multicolumn{1}{c|}{4} & 1               & 4               \\ \midrule
ReDiff-Base     & 71.31 & 71.67                  & 51.73 & 51.83                  & 58.04  & 55.22                  & -69.17          & -73.17              \\
Online training round 1        & \textbf{76.74} & \textbf{75.85}                  & 55.07 & \textbf{54.08}                  & \textbf{63.29}  & \textbf{60.87}                 & \textbf{-51.50}          & \textbf{-72.67}             \\
Online training round 2            & 76.10 & 74.99                  & \textbf{55.20} & \textbf{54.08}                  & 62.24  & 60.46                  & -56.17          & -72.83          \\
\bottomrule
\end{tabular}
\end{adjustbox}
\vspace{-0.2cm}
\end{table}

\subsection{Ablation Studies}

\textbf{Impact of Each Training Stage.} 
In Table \ref{tab:train stage}, we analyze the individual contributions of our two training stages. Both Stage I (foundational revision) and Stage II (self-correction) independently improve the model's performance and stability over the LLaDA-V baseline. Furthermore, the most significant gains are achieved when both stages are combined. Notably, Stage II alone provides a more substantial boost than Stage I, confirming that teaching the model to learn from its own intrinsic errors is a highly effective refinement strategy. After Stage I training, the model exhibits stable parallel generation performance. For example, as the speed increases from 1 to 4 tokens/step, CLAIR improves from 71.31 to 71.67, and CapArena changes from -69.17 to -73.17.  The combination of the two stages yields a synergistic effect, with Stage I providing a foundational revision ability that is further amplified by Stage II, leading to large improvements in metrics like Factuality (+5.25) and CapArena (+17.67).

\textbf{Analysis of Foundational Revision Training.} 
As shown in Table \ref{tab:stage1}, we investigate different settings for the stage I training. We find that revising syntactic errors primarily boosts overall quality (CLAIR) and Coverage, while also enhancing stability during parallel generation. Conversely, training on hallucination revision exhibits higher Factuality. Combining both error types allows our model to achieve the best overall performance. We also compare dynamic probability for random token replacement in the fourth line, where the dynamic rate is correlated with the noise level t (using t as replacement probability, when t $\textless$ 0.1). The results indicate that our fixed replacement rate yields better overall performance.

\textbf{Impact of Online Self-Correction Training Rounds.} 
In Table \ref{tab:stage-2}, we examine the effect of iteration of the stage II training. The results show that while the first round of self-correction provides a substantial performance boost over the ReDiff-Base model, subsequent rounds of training on newly generated data do not yield further significant improvements across most metrics.

\begin{figure*}[h]
	\centering
	\includegraphics[width=\linewidth]{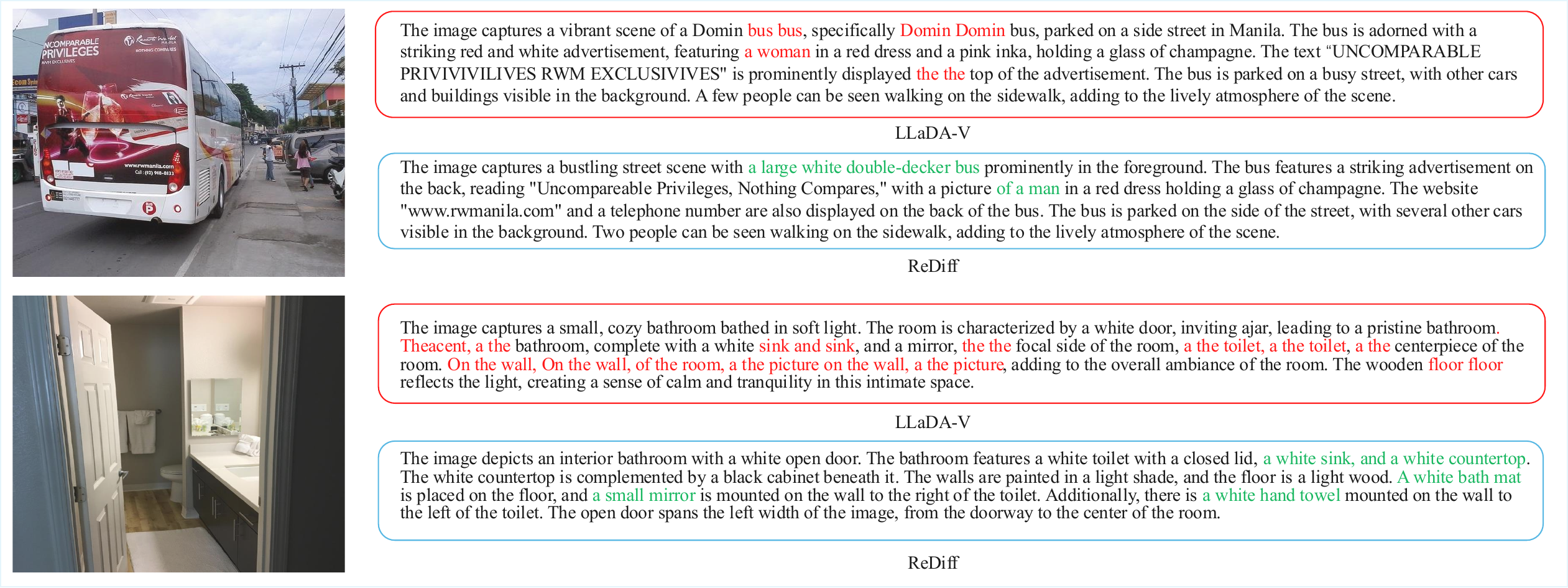}
	\caption{Cases comparison between LLaDA-V and our ReDiff under 4 tokens/step inference speed. ReDiff demonstrates superior fluency and accuracy in its generated captions.}
	\label{fig:vis}
\end{figure*}

\begin{figure*}[h!]
	\centering
	\includegraphics[width=0.95\linewidth]{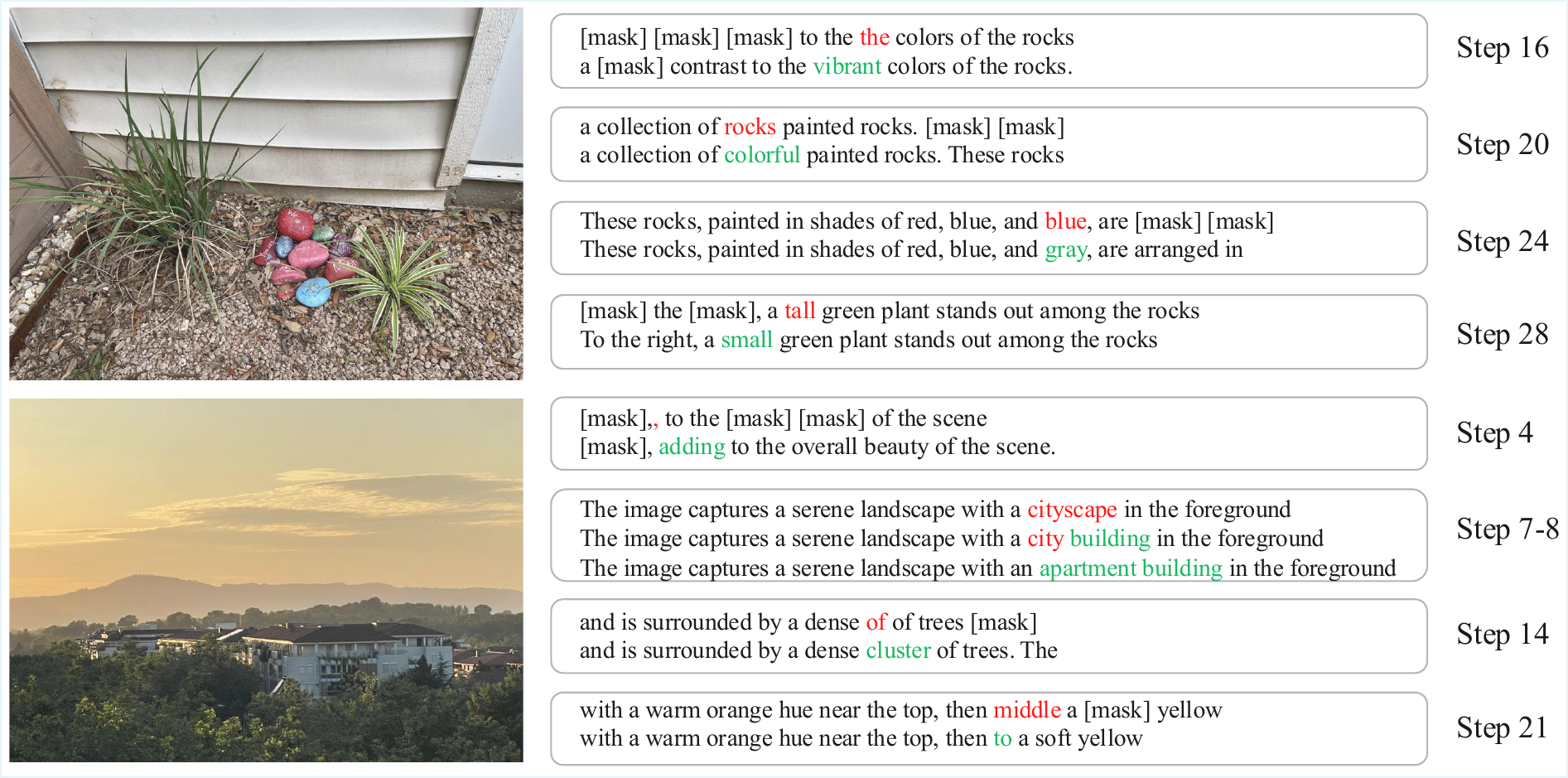}
	\caption{Refinement process of ReDiff at different inference step. \textcolor{red}{Red tokens} indicate the errors produced during generation, while \textcolor{green}{green tokens} mean the corresponding refined results.}
	\label{fig:refine process}
\end{figure*}

\begin{figure*}[h]
	\centering
	\includegraphics[width=\linewidth]{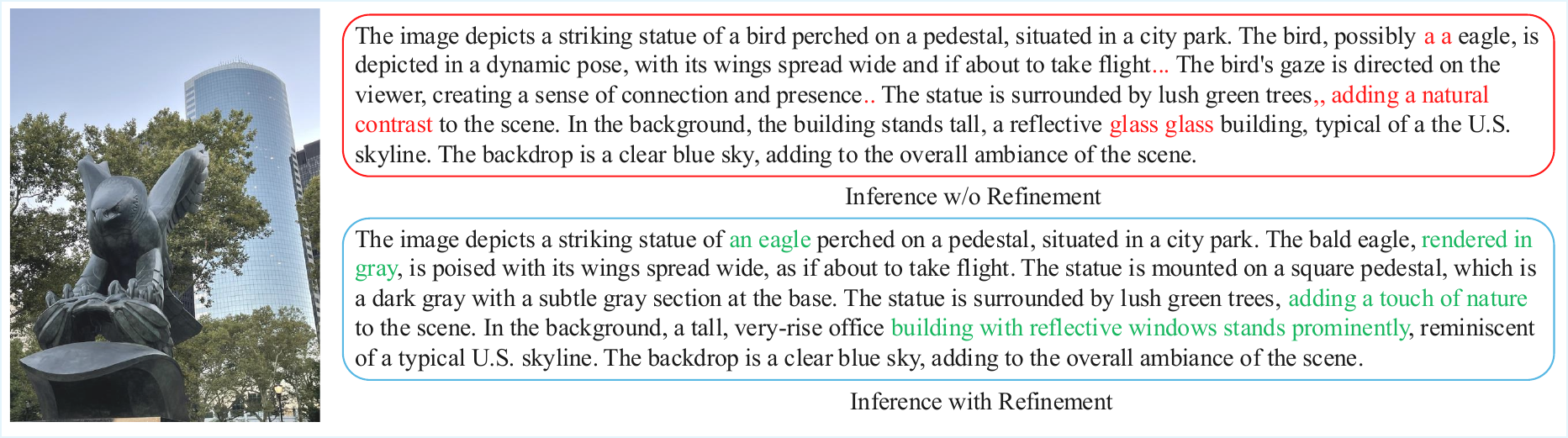}
	\caption{Generation results of ReDiff with or w/o refinement during inference.}
	\label{fig:refinement}
\end{figure*}

\begin{figure*}[h]
	\centering
	\includegraphics[width=\linewidth]{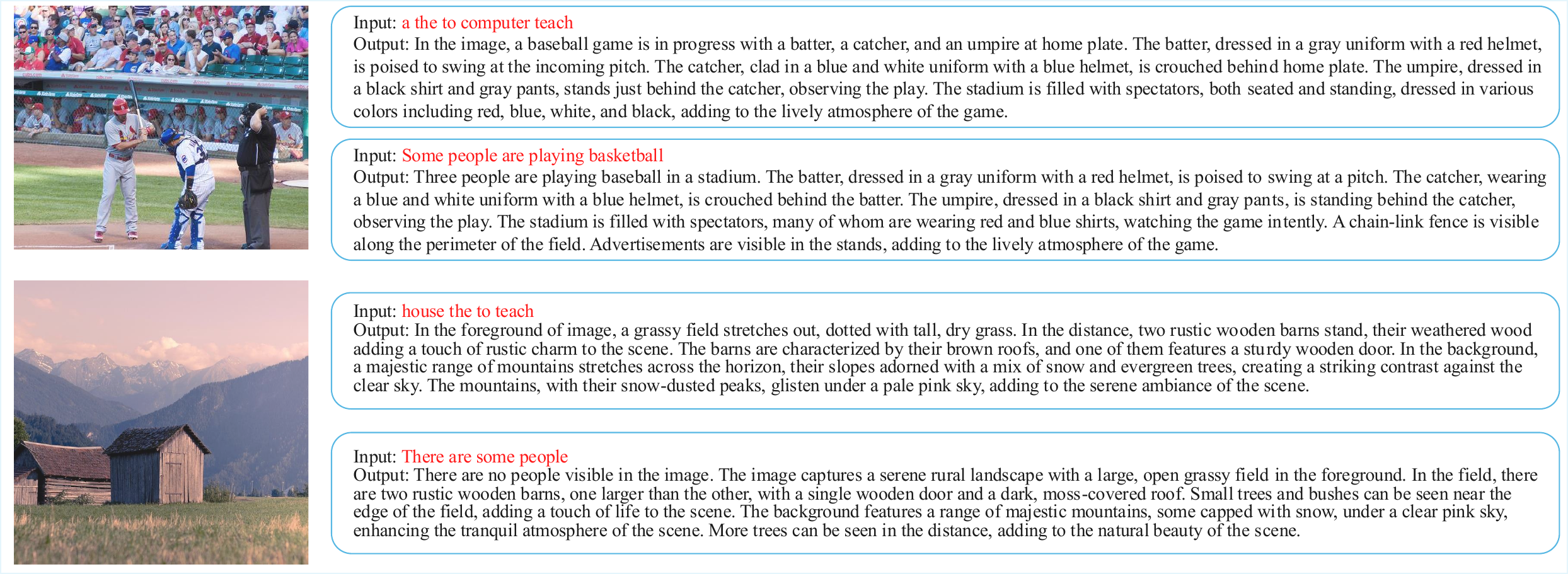}
	\caption{ReDiff can revise wrong input answers.}
	\label{fig:wrong}
\end{figure*}

\subsection{Qualitative Analysis}

We provide qualitative examples to visually demonstrate how the refinement during inference produces more accurate and fluent results, thereby improving the stability of parallel generation.

In Figure \ref{fig:vis}, we compare the parallel-generated captions from ReDiff and LLaDA-V. The baseline's output suffers from token repetition (``bus", ``the"), grammatical errors, and hallucinations (e.g., misidentifying a person on a bus advertisement as ``a woman"). In contrast, our model's output is fluent, coherent, and factually grounded. In the second example, our model accurately describes all key elements in the scene, whereas the baseline's output is chaotic and omits significant details. More comparison cases can be found in Appendix \ref{app:vis}.

Figure \ref{fig:refine process} visualizes the token-level changes during a 32-step generation process. It clearly shows the model simultaneously unmasking new tokens and refining previously generated ones. For instance, in the first example, the model refines the erroneous phrase ``rocks painted rocks" into ``colorful painted rocks" at step 20. At step 28, it corrects ``a tall green plant" to ``a small green plant" to better match the visual content. 

Figure \ref{fig:refinement} showcases comparison of inference with and without the refinement, showing that the refinement is critical for achieving high-quality outputs. 
If ReDiff inferences without the refinement, errors tend to accumulate, such as repeated words or symbols and incoherent sentences, ultimately degrading the quality of the caption. This highlights the importance of the model’s refinement capability. 

Beyond correcting the model's own errors during generation, ReDiff also demonstrates a powerful, generalizable ability to revise disturbing inputs. As shown in Figure \ref{fig:wrong}, we provide the model with an image and a user-provided caption containing either syntactic chaos or a factual hallucination. In both cases, our model successfully corrects the initial erroneous text and proceeds to generate a coherent and accurate completion, highlighting the strong revision ability of ReDiff.


\section{Conclusion}

In this work, we addressed the critical challenge of error cascades that hampers the performance of vision-language diffusion models, particularly in efficient parallel generation scenarios. We proposed a paradigm shift from passive denoising to active refining by introducing ReDiff, a novel framework centered on a mistake-driven, online self-correction loop. This approach teaches the model to learn from its own characteristic errors, endowing it with the ability to revisit and refine its generated output. Our extensive experiments validate that this method not only achieves state-of-the-art performance but, more importantly, demonstrates far superior stability and factual accuracy in challenging few-step generation regimes where traditional denoising models catastrophically fail. By effectively breaking the error cascade, our work presents a promising path toward developing more robust, efficient, and controllable generative systems.
\clearpage
\bibliography{iclr2026_conference}
\bibliographystyle{iclr2026_conference}

\clearpage
\newpage
\appendix

\section{More Visualization}
\label{app:vis}

\begin{figure*}[hb]
	\centering
	\includegraphics[width=\linewidth]{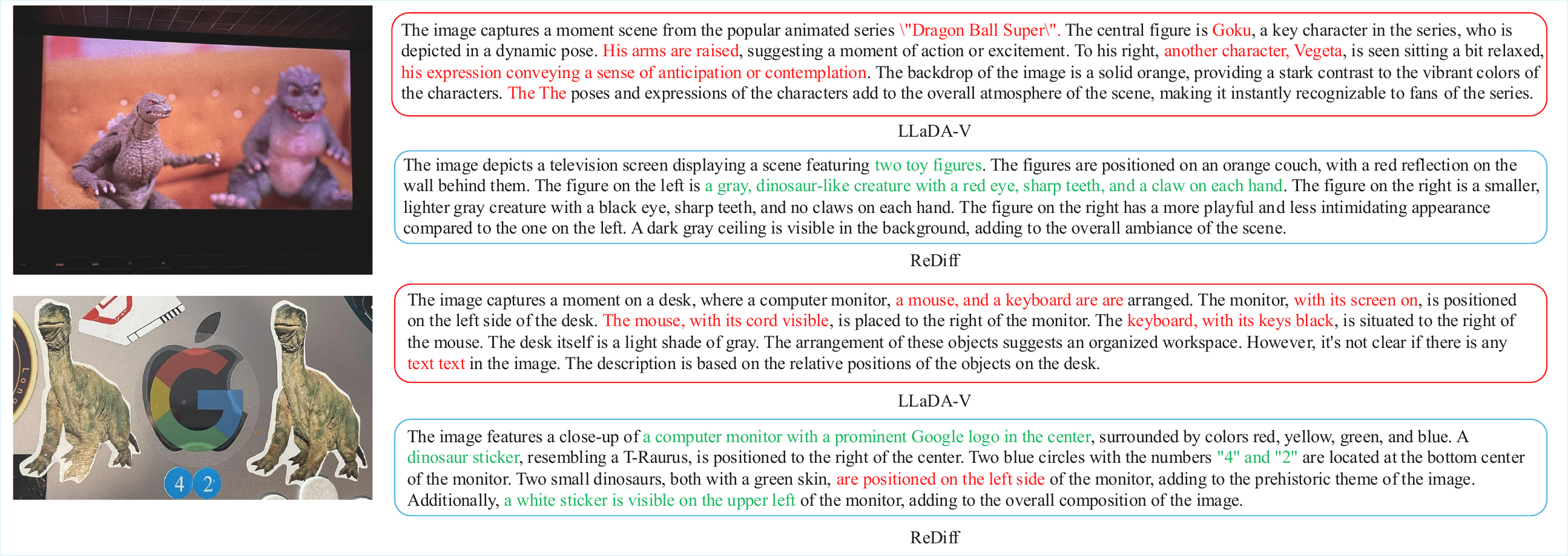}
	\caption{Cases comparison between LLaDA-V and our ReDiff under 2 tokens/step inference speed.}
	\label{fig:case2}
\end{figure*}

\begin{figure*}[hb]
	\centering
	\includegraphics[width=\linewidth]{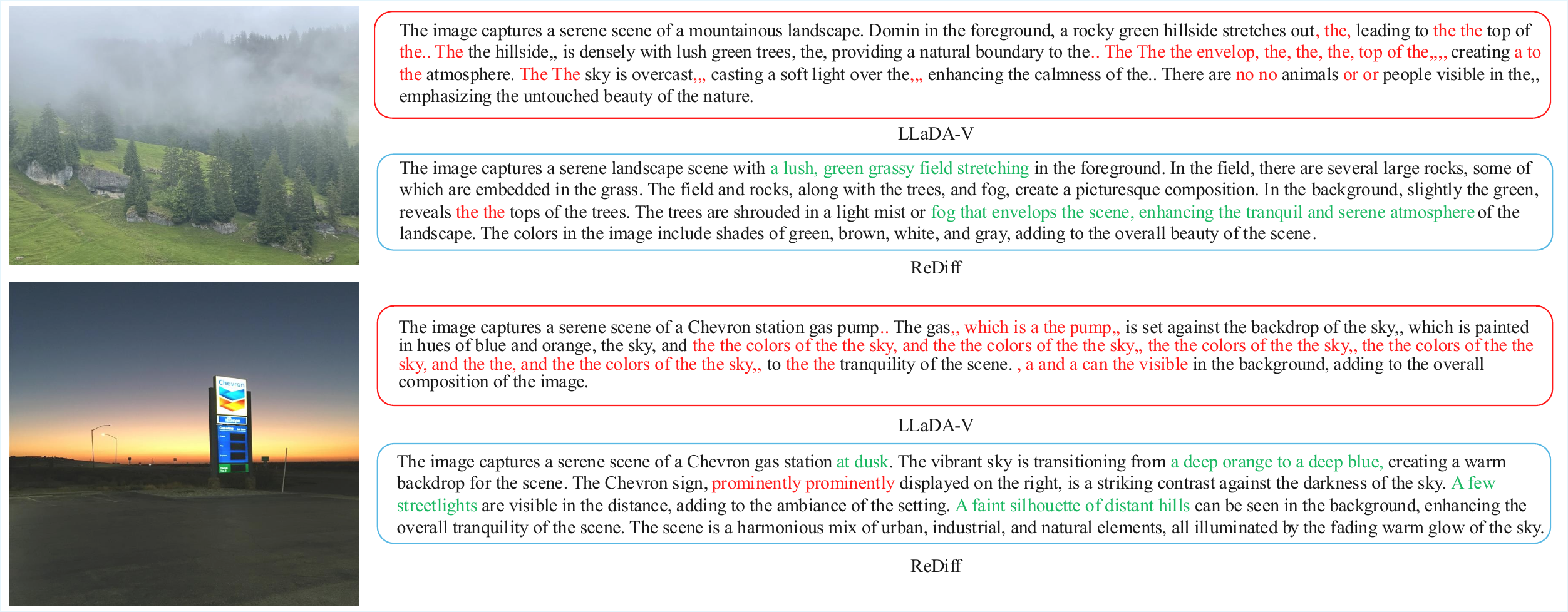}
	\caption{Cases comparison between LLaDA-V and our ReDiff under 8 tokens/step inference speed.}
	\label{fig:case8}
\end{figure*}

Figure \ref{fig:case2} and Figure \ref{fig:case8} show the generation results of ReDiff and LLaDA-V under different inference steps. In the 2 tokens/step scenario, LLaDA-V outputs a great deal of hallucinated content, such as ``Goku" and ``Vegeta" in the first case, and a mouse and keyboard in the second. This occurs because an initial hallucination can affect subsequent generation, leading to error cascades. In contrast, our ReDiff method produces captions that are consistent with the image content. In the cases of 8 tokens/step, the results of our model are more fluent and have less grammer errors.

\section{Prompt for Stage II data Construction}
\label{app:prompt}

In online self-correction learning loop, we use ReDiff-Base to generate caption drafts. Then the image, the draft, and ground truth caption are fed to o4-mini to detect and revise errors. The prompt for o4-mini is shown in Table \ref{tab:o4 prompt}.

\begin{table*}[t]\centering
\label{tab:o4 prompt}
\small
\begin{minipage}{1.0\textwidth}\vspace{0mm}    \centering
\caption{The prompt for o4-mini to revise model's drafts.}
\label{tab:o4 prompt}
\begin{tcolorbox} 
    \centering
    \begin{tabular}{p{0.97\textwidth} c}

\# ROLE: Hallucination detect and revise Assistant  & \\
\\

\#\# PERSONA:& \\

You are an AI assistant specialized in hallucination revision. You integrate information from image, question and ground truth answer, to analyze and judge whether the prediction from other models is right or not. If the prediction is wrong, you need to revise the hallucination and errors in the prediction.  & \\
\\
\#\# INPUT CONTEXT:& \\

You will receive the following:& \\
1.  **Image:** An image.& \\
2.  **Question:** A user's question about the image.& \\
3.  **Answer:** The right answer to the question.& \\
4.  **Prediction** The answer from our model. & \\
\\

\#\# TASK:& \\
Your primary task is to judge if the prediction is right according to the image and ground truth answer. If the prediction is not right, detect hallucination and wrong parts, then revise them.& \\
* The prediction must be consistent to the image, detect all hallucination and errors.& \\
* For the words containing hallucination, you need to replace them with right words, which have same token number with original prediction. Make the original prediction correct with as few modifications as possible. & \\
* The answer may contain some chaos in grammar expression, such as repetition, incoherence, etc. You should also replace erroneous parts with fragments of identical token counts.& \\
\\

\#\# GUIDELINES \& CONSTRAINTS:& \\
1.  The prediction doesn't have to be identical to the reference answer; as long as it correctly answers the question, it's acceptable. The GT (ground truth) serves merely as a reference. Focus primarily on checking whether there are hallucination issues in the prediction that contradict the image content.& \\
2.  If the prediction is correct, output only `right'. & \\
3.  If the prediction contains hallucinations or errors, output a JSON-formatted string containing multiple pairs of phrases. Each pair should consist of the original erroneous phrase segment and its corrected counterpart. & \\
4.  Modifications should be localized to the minimal necessary extent, typically targeting short multi-word segments. & \\
5.  For each pair, ensure the tokenized length of the original and modified segments remains identical. The semantics of replacement words must be inconsistent to the original segment.& \\
6.  The original segment should be unique within the prediction to facilitate error localization by users.& \\
\\

\#\# OUTPUT FORMAT:& \\
1.  If the prediction is right, output only `right'. & \\
2.  If the prediction has errors, provide the output as a single JSON object, which is a list containing multiple dictionaries with the following keys:& \\
    * `org': (String) The hallucination or error segment in original prediction.& \\
    * `target': (String) The right segment to replace the wrong part in prediction.& \\
\\
\#\# EXAMPLE:

* **Image:** [Description: A man in front of a white trunk.]& \\
* **Question:** ``What might the man in the suit be doing?"& \\
* **Answer:** ``The man dressed in business attire leaning on the white truck could be associated with the business related to the truck ..."& \\
* **Prediction:**``The man is leaning on a pink trunk, and ..."& \\
* **Expected Output:**& \\
    ```json 
    [
        {
        ``org": ``a pink truck",
        ``target": ``a white trunk"
        }
    ] & \\
\\
    \end{tabular}
\end{tcolorbox}
    
\end{minipage}
\end{table*}

\section{Ethics Statement}

All datasets and models used in this study are publicly available and open-source. No proprietary, private, or personally identifiable information was collected or used. The images employed are either natural scenes or normal human activities, without any violent, explicit, or otherwise harmful content. Therefore, the research meets relevant considerations regarding privacy, ethics and copyright. 

\end{document}